\documentclass[letterpaper, 10 pt, conference]{ieeeconf}  % Comment this line out if you need a4paper

\IEEEoverridecommandlockouts                              % This command is only needed if 
\usepackage{graphics} % for pdf, bitmapped graphics files
\usepackage{graphicx,import}
\usepackage{subfigure}
\usepackage{amsmath,amssymb,amsfonts} % assumes amsmath package installed

\usepackage{tikz}                      %language for producing vector graphics from a geometric/algebraic description
\usepackage{tikzscale}
\usepackage{relsize}
\usepackage{pgfplots}                  %draws high-quality function plots in normal or logarithmic scaling
\pgfplotsset{compat=newest}
\usepgfplotslibrary{external}
\usepackage{cite}
\usepackage{textcomp}

\usepackage{marvosym} % package for \kutline (---)

\newcommand\copyrighttext{%
    \footnotesize \copyright~2018 IEEE. Personal use of this material is permitted. Permission from IEEE must be obtained for all other uses, in any current or future media, including reprinting/republishing this material for advertising or promotional purposes, creating new collective works, for resale or redistribution to servers or lists, or reuse of any copyrighted component of this work in other works. DOI: 10.1109/ITSC.2018.8569701}%
\newcommand\copyrightnotice{%
    \begin{tikzpicture}[remember picture,overlay]
        \node[anchor=south,yshift=10pt] at (current page.south) {\fbox{\parbox{\dimexpr\textwidth-\fboxsep-\fboxrule\relax}{\copyrighttext}}};
    \end{tikzpicture}%
}

\title{\LARGE \bf
Multi-Object Tracking with Interacting Vehicles \\and Road Map Information
}

\author{Andreas Danzer, Fabian Gies and Klaus Dietmayer% <-this % stops a space
\thanks{A. Danzer, F. Gies, and K. Dietmayer are with the Institute of Measurement, Control and Microtechnology, Ulm University, 89081 Ulm, Germany
        {\tt\small firstname.lastname@uni-ulm.de}}%
}

\graphicspath{{img/}}

\newlength\figH
\newlength\figW
\def\hlinewd#1{%
    \noalign{\ifnum0=`}\fi\hrule \@height #1 %
    \futurelet\reserved@a\@xhline} 

\begin{document}

\maketitle
\copyrightnotice
\thispagestyle{empty}
\pagestyle{empty}

%%%%%%%%%%%%%%%%%%%%%%%%%%%%%%%%%%%%%%%%%%%%%%%%%%%%%%%%%%%%%%%%%%%%%%%%%%%%%%%%
\begin{abstract}
    In many applications, tracking of multiple objects is crucial for a perception of the current environment. 
Most of the present multi-object tracking algorithms assume that objects move independently regarding other dynamic objects as well as the static environment.
Since in many traffic situations objects interact with each other and in addition there are restrictions due to drivable areas, the assumption of an independent object motion is not fulfilled. 
This paper proposes an approach adapting a multi-object tracking system to model interaction between vehicles, and the current road geometry. 
Therefore, the prediction step of a Labeled Multi-Bernoulli filter is extended to facilitate modeling interaction between objects using the Intelligent Driver Model. 
Furthermore, to consider road map information, an approximation of a highly precise road map is used. 
The results show that in scenarios where the assumption of a standard motion model is violated, the tracking system adapted with the proposed method achieves higher accuracy and robustness in its track estimations.
\end{abstract}

%%%%%%%%%%%%%%%%%%%%%%%%%%%%%%%%%%%%%%%%%%%%%%%%%%%%%%%%%%%%%%%%%%%%%%%%%%%%%%%%
\section{Introduction}
\label{sec:introduction}
Tracking of multiple objects, also known as multi-object tracking, is a basic prerequisite in many applications such as traffic surveillance, advanced driver assistance systems and autonomous driving. 
The aim of multi-object tracking algorithms is jointly estimating the number of objects and their individual states using a sequence of noisy measurements. 
Three common approaches are Joint Probabilistic Data Association (JPDA) \cite{Bar-Shalom1988}, Multiple Hypotheses Tracking (MHT) \cite{Reid1979} and the multi-object Bayes filter, which is based on the Random Finite Set (RFS) framework \cite{Mahler2007}. 
The $\delta$-Generalized Labeled Multi-Bernoulli ($\delta$-GLMB) filter \cite{Vo2013} is the first analytic implementation of the multi-object Bayes filter using so called labeled RFSs. For the reason that the computational complexity of the $\delta$-GLMB filter increases exponentially with the number of objects, this filter only is suitable for scenarios with a small number of objects. 
The Labeled Multi-Bernoulli (LMB) filter \cite{Reuter2014} approximates the $\delta$-GLMB filter in an efficient way representing the posterior density by an LMB distribution, however this approximation results in a loss of information.
%Although this approximation results in a loss of information, the LMB filter achieves almost the same accuracy as the $\delta$-GLMB filter \cite{Reuter2014b}. 
In contrast to the LMB filter, the $\delta$-GLMB filter facilitates modeling dependencies between objects using a $\delta$-GLMB distribution with multiple hypotheses. 
Since the LMB filter comprises various hypotheses within the spatial distribution of the estimated object \cite{Reuter2014b}, the dependence on other objects is lost. 
However, in today's traffic the motion of objects is based on interaction between dynamic objects among each other, as well as regarding the static environment. 
There are two situations using the example of vehicles in traffic, in which interaction is of great importance. 
Firstly, in city traffic with many vehicles at the same time, the velocity of a vehicle depends on the velocity of the vehicle ahead. 
Secondly, the vehicle's direction of motion strongly depends on the road map respectively the road geometry, e.g. curves of the road, roundabouts or intersections. 
Since, the standard LMB filter assumes that all objects move independently, it does not consider interaction within the current environment of objects, neither other estimated objects nor road geometry.
As shown in the two examples presented, the model assumptions of the LMB filter do not correspond to reality. 
An improvement of the tracking results can be achieved by considering interaction between dynamic objects as well as the static environment.

There are several approaches considering other objects and the environment within a multi-object tracking algorithm. 
In \cite{Luber2010}, physical and social constraints of the environment are integrated into MHT for pedestrians. 
For this purpose, the prediction of a pedestrian is adapted to the existence of other people, obstacles and physical constraints using the social force model (SFM) \cite{Helbing1995} combined with a Kalman filter based tracker. 
In \cite{Feng2017}, the SFM is used to describe interaction between multiple persons and to adapt the likelihood within the prediction step of a probability hypotheses density (PHD) filter \cite{Mahler2003} using a Markov chain Monte Carlo implementation. 
Both approaches only model interaction between pedestrians, since the SFM is developed to describe motion of pedestrians. 
Considering physical road constraints and interaction between vehicles using a force-based dynamic model, the domain knowledge-aided moving horizon estimation method combined with an MHT structure is proposed in \cite{Ding2017}. 
Therefore, the SFM is modified modeling the dynamics of vehicles and integrated into a vehicle tracking system. 
A critical aspect is the SFM models interaction with virtual forces what makes it hard using for describing vehicle dynamics with a physical interpretation. 
In consequence, the Generalized Force Model (GFM) \cite{Helbing1998} is introduced for modeling the motion of vehicles, respectively the dynamics of interacting vehicles. 
The GFM is a car-following model describing the dynamics of a vehicle regarding to a vehicle ahead in traffic flow. 
There are several other car-following models in literature, e.g. the linear Helly model \cite{Helly1959} and the Intelligent Driver Model (IDM) \cite{Treiber2000}. 
In \cite{Song2017}, a multi-vehicle tracking system considering road map information and vehicles interaction using a car-following model is presented. 
For this purpose, it is assumed that the movement of vehicles is limited on single lane roads without any intersections. 
Further, the Helly model is integrated into a framework using the standard Kalman filter \cite{Kalman1960}, which only is applicable for linear vehicle motion.
The tracking system represents the estimation of vehicles in road coordinates and measurements are generated from a simulated GPS sensor. 
Additionally, measurements are validated before data association using the road map information.

This contribution proposes another approach considering information about the road geometry and interaction between estimated tracks representing vehicles. 
Modeling track dependencies, a car-following model, here the IDM, is integrated into a Gaussian Mixture (GM) LMB filter implementation \cite{Reuter2014b} with an underlying Unscented Kalman Filter (UKF) \cite{Julier2004}.
Using an UKF facilitates dealing with nonlinear filter equations, so the motion of vehicles can be described more realistic.  
Further, the information of a highly precise digital map is transformed into a suitable approximation to quickly incorporate the road map information into the GM-LMB filtering steps. 
In contrast to the method presented in \cite{Song2017}, this approach is not limited to roads without intersections. 
Another important aspect is, the presented approach is integrated into the real-time environment perception system using in the autonomous vehicle of Ulm University \cite{Kunz2015}. 
Since real world data is used, the results of this contribution are representative.

In many tracking application, a Constant Turn Rate and Velocity (CTRV) model \cite{Schubert2008} describes the vehicle dynamics. 
In reality this assumption of motion, due to interaction between objects among each other and the road geometry, is not fulfilled, i.e. velocity and turn rate are not constant. 
For this reason, this paper presents the integration of interaction between objects using the IDM and road geometry information using a new approach in the prediction step of the GM-LMB filter. 
The main idea is to correct the wrong assumption of the dynamic model, i.e. velocity and turn rate, adapting the prediction step of a GM-LMB filter. 

The paper is organized as follows. 
Section \ref{sec:background} recaps the basics of labeled RFSs and the LMB filter. Further, the IDM is introduced. 
Section \ref{sec:interactedprediction} describes the GM-LMB filter first.
Subsequently, the calculation of road map information, as well as the modeling of interaction between objects and considering road map information in the GM-LMB filter is presented. 
Section \ref{sec:results} presents results using realistic scenarios with real data.
Finally, in Section \ref{sec:conclusion} concluding remarks are given.

\section{Background}
\label{sec:background}
This section recaps the principle of multi-object tracking using labeled multi-Bernoulli Random Finite Sets and provides the basics of the Labeled Multi-Bernoulli filter.
Also, the Intelligent Driver Model (IDM) later used for interaction between objects is reviewed.

\subsection{Labeled Multi-Bernoulli RFS}

An RFS \cite{Mahler2007} is a finite-set-valued random variable containing a random number of unordered points which are also random.
The multi-object state is represented by the RFS 
$\text{X} = \{\mathrm{x}^{(1)},\dots,\mathrm{x}^{(N)}\} \subset \mathbb{X}$ with finite single-target state vectors $\mathrm{x}^{(i)} \in \mathbb{X}$ and the state space $\mathbb{X}$.
%Further, the RFS $\text{Z} = \{\mathrm{z}^{(1)},\dots,\mathrm{z}^{(M)}\} \subset \mathbb{Z}$ represents the multi-object observations with a random measurement $\mathrm{z}^{(i)} \in \mathbb{Z}$, where $\mathbb{Z}$ is the measurement space.

Either, a Bernoulli RFS is empty with a probability $1-r$, or is a singleton with probability $r$ and spatial distribution $p$.
The probability density is
\begin{equation}
\pi (\text{X}) = 
\begin{cases}
1 - r &\text{X} = \emptyset, \\
r \cdot p(\mathrm{x}) &\text{X} = \{\mathrm{x}\}.
\end{cases}
\end{equation}
A multi-Bernoulli RFS comprises multiple independent Bernoulli RFSs $\text{X}^{(i)}$, thus, $\text{X} = \bigcup_{i=1}^{M} \text{X}^{(i)}$. 
%The parameter set $\{(r^{(i)},p^{(i)})\}_{i=1}^{M}$ completely describes such an RFS, whereby $r^{(i)}$ is the existence probability and $p^{(i)}$ the spatial distribution.

In a scenario with multiple objects, a key objective is to estimate the current state as well as the identity of an object. 
Therefore, \cite{Vo2013} introduces the class of labeled RFSs. In a labeled RFS, for each state vector $\mathrm{x} \in \mathbb{X}$ a label $\ell \in \mathbb{L}$ is attached, where $\mathbb{L}$ is a finite label space.

A labeled multi-Bernoulli (LMB) RFS is completely defined by the parameter set
$\{(r^{(\ell)},p^{(\ell)})\}_{\ell \in \mathbb{L}}$, whereby $r^{(\ell)}$ is the existence probability and $p^{(\ell)}$ the spatial distribution.
According to \cite{Reuter2014} the corresponding density is
\begin{equation}
\boldsymbol \pi(\textbf{X}) = \delta_{\vert \textbf{X} \vert}(\vert \mathcal{L}(\textbf{X}) \vert) \, w(\mathcal{L}(\textbf{X})) p^{\textbf{X}},
\label{eq:lmbdensity}
\end{equation}
with 
\begin{align}
w(L) &= \prod\limits_{i \in \mathbb{L}}\left(1-r^{(i)}\right) \prod\limits_{\ell \in L}\dfrac{1_{\mathbb{L}}(\ell)r^{(\ell)}}{1-r^{(\ell)}}, \label{eq:lmb_weight}\\
p(\mathrm{x},\ell) &= p^{(\ell)}(\mathrm{x}), 
\end{align}
where  $\delta_{\vert \textbf{X} \vert}(\vert \mathcal{L}(\textbf{X}) \vert)$ ensures distinct labels $\ell$ of a realization.
The projection in (\ref{eq:lmbdensity}) maps labeled state vectors to labels using  ${\mathcal{L}((\mathrm{x},\ell)) = \ell}$ with $\mathcal{L} : \mathbb{X} \times \mathbb{L} \rightarrow \mathbb{L}$.

\subsection{Labeled Multi-Bernoulli Filter}
\label{ssec:lmbfilter}

The Labeled Multi-Bernoulli (LMB) filter \cite{Reuter2014} is an accurate and fast multi-object tracking filter using RFSs.
An advantage of the LMB filter is modeling uncertainty in data association implicitly. Considering that, the spatial distribution of a track comprises the association of tracks to multiple measurements.

An LMB RFS is a conjugate prior with respect to the prediction equations. The multi-object prediction of a multi-object posterior LMB RFS with parameter set ${\boldsymbol \pi = \{(r^{(\ell)},p^{(\ell)})\}_{\ell \in \mathbb{L}}}$ on $\mathbb{X} \times \mathbb{L}$ and an LMB birth density $\boldsymbol \pi_B = \{(r_B^{(\ell)},p_B^{(\ell)})\}_{\ell \in \mathbb{B}}$ on $\mathbb{X} \times \mathbb{B}$, is again an LMB RFS defined on the state space $\mathbb{X}$ and finite label space ${\mathbb{L}_+ = \mathbb{B} \cup \mathbb{L}}$.
The prediction is given by
\begin{equation}
\boldsymbol \pi_+\left(\textbf{X}\right) = \left\{\left( r_{+,S}^{(\ell)},p_{+,S}^{(\ell)} \right)\right\}_{\ell \in \mathbb{L}} \cup \left\{\left( r_{B}^{(\ell)},p_{B}^{(\ell)} \right)\right\}_{\ell \in \mathbb{B}},
\label{eq:lmb_prediction}
\end{equation}
where
\begin{align}
r_{+,S}^{(\ell)} & = \eta_S(\ell) r^{(\ell)},\\
p_{+,S}^{(\ell)} & = \dfrac{\langle p_S(\cdot,\ell) f(\mathrm{x} \vert \cdot,\ell), p(\cdot,\ell) \rangle}{\eta_S(\ell)},\\
\eta_S(\ell) &= \int \langle p_S(\cdot,\ell) f(\mathrm{x} \vert \cdot,\ell),p(\cdot,\ell) \rangle \mathrm{d}\mathrm{x}.
\label{eq:lmb_prediction_end}
\end{align}
Here, $p_S(\cdot,\ell)$ describes the survival probability and $f(\mathrm{x} \vert \cdot,\ell)$ the target transition density for a single track $\ell$.
Further, $\eta_S(\ell)$ is a normalization constant.
For a detailed explanation and the equations of the LMB filter update step see \cite{Reuter2014b}.

\subsection{The Intelligent Driver Model}
\label{ssec:idm}

The Intelligent Driver Model (IDM) \cite{Treiber2000} is a car-following model which describes the longitudinal dynamics of a vehicle following a leading vehicle along a single lane. 
The IDM calculates the acceleration of a vehicle $\alpha$ depending on its velocity $v_{\alpha}$, the distance $s_{\alpha}$ and the velocity difference $\Delta v_{\alpha}$ to the leading vehicle $\alpha-1$. The acceleration is given by $\dot{v}_{\alpha} = \dot{v}_{\alpha}^{\text{free}} + \dot{v}_{\alpha}^{\text{interaction}}$, where $\dot{v}_{\alpha}^{\text{free}}$ describe the acceleration of vehicle $\alpha$ without leading vehicle $\alpha-1$ achieving a desired velocity, and $\dot{v}_{\alpha}^{\text{interaction}}$ the deceleration due to the leading vehicle. 
Since in tracking application, a desired velocity is not known, in the following the acceleration $\dot{v}_{\alpha}^{\text{free}}$ is negligible. 
The interactive deceleration compares the current distance $s_{\alpha}$ with the desired distance  $s^{\ast}(v_{\alpha},\Delta v_{\alpha})$ and is given by
\begin{equation}
    \dot{v}_{\alpha}^{\text{interaction}} = -a \left( \dfrac{s^{\ast}(v_{\alpha},\Delta v_{\alpha} )}{s_{\alpha}}\right)^2,
    \label{eq:vidm}
\end{equation}
where $a$ is the maximum acceleration.
The desired distance is
\begin{equation}
    s^{\ast}(v_{\alpha},\Delta v_{\alpha}) = s_0 + v_{\alpha}T + \dfrac{v_{\alpha} \: \Delta v_{\alpha}}{2\sqrt{ab}},
    \label{eq:sidm}
\end{equation}
where $s_0$ is the minimum distance, $v_{\alpha}T$ a safety distance depending on the velocity $v_{\alpha}$ and time gap $T$, and an intelligent deceleration strategy using the maximum acceleration $a$ and the comfortable deceleration delay $b$.

\section{LMB Filter modeling interaction between vehicles among each other and road geometry}
\label{sec:interactedprediction}
As shown in Section \ref{ssec:lmbfilter}, an object consists of a label, an existence probability and the spatial distribution. The prediction step propagates the existence probability and the spatial distribution. The prediction of the spatial distribution depends on the modeling of the vehicle's dynamics. This contribution proposes an approach to incorporate interaction between vehicles with each other and with the road geometry during the prediction step.

\subsection{The Gaussian Mixture Labeled Multi-Bernoulli Filter}
\label{ssec:gmlmbfilter}

The LMB filter is realized using a Gaussian Mixture (GM) implementation introduced in \cite{Reuter2014b}.
For this purpose, a mixture of Gaussian distributions
\begin{equation}
p^{(\ell)}(\mathrm{x}) = \sum_{j = 1}^{J^{(\ell)}} w^{(\ell,j)} \mathcal{N} \left(\mathrm{x}; \hat{\mathrm{x}}^{(\ell, j)},\mathrm{P}^{(\ell, j)}\right),
\label{eq:gaussian_mixture}
\end{equation}
represents the posterior probability densities $p^{(\ell)}$ of all labeled Bernoulli tracks $\ell \in \mathbb{L}$.
In (\ref{eq:gaussian_mixture}), $\hat{\mathrm{x}}^{(\ell, j)}$ is the mean value of respective Gaussian component and $\mathrm{P}^{(\ell, j)}$ the corresponding error covariance estimation, where each Gaussian component describes one possible state of track $\ell$. 
As described in \cite{Vo2009}, the weights $w^{(\ell,j)}$ of the components are normalized to one.

In the case, that each object follows a linear Gaussian process model, the prediction step using standard Kalman filter equations is shown in \cite{Reuter2014b}.

Since in realistic scenarios, the object dynamics is nonlinear, the standard Kalman filter is not sufficient. 
The prediction step to slightly nonlinear motion models may be realized with an Extended Kalman Filter (EKF) \cite{Bar-Shalom1988} or an Unscented Kalman Filter (UKF) \cite{Julier2004} implementation. 
Here the UKF implementation for prediction and update of the posterior probability densities $p^{(\ell)}$ is used.
The UKF samples several points, also known as sigma points, according to the principles of the unscented transform. 
The predicted and updated densities are given by the propagation of these points through the nonlinear motion and measurements transformations. 
The UKF approach facilitates adapting individual sigma points using interaction between objects and road map information.
This has the major advantage that not only the mean of the object's state is adapted, but also implicitly its covariance, since the covariance matrix is calculated using each sigma point.
For more details and corresponding equations see \cite{Julier2004}.

Obtaining a closed form prediction of individual GM components, the unscented transform is applied to propagate the first and second moment of the predicted densities through the nonlinear motion model.
So, for each GM component (\ref{eq:gaussian_mixture}), i.e. each Gaussian distribution $\mathcal{N} \left(\mathrm{x}_{k-1}; \hat{\mathrm{x}}^{(\ell, j)}_{k-1}, \mathrm{P}^{(\ell, j)}_{k-1}\right)$, a set of sigma points is generated.
At time $k-1$, the set of sigma points per mixture component consists of $2n+1$ sigma points with augmented mean and covariance
\begin{align}
\mu^{(\ell, j)}_{k} &= \left[ \hat{\mathrm{x}}^{(\ell, j)}_{k-1}, \; 0^\text{T}\right],\\
\Sigma^{(\ell, j)}_{k} &= \text{diag}\left(\mathrm{P}^{(\ell, j)}_{k-1}, \; \mathrm{Q}_{k-1}\right),
\label{eq:augmentedmean}
\end{align}
where $\hat{\mathrm{x}}^{(\ell, j)}_{k-1}$, $\mathrm{P}^{(\ell, j)}_{k-1}$ is the mean with its corresponding covariance and $\mathrm{Q}_{k-1}$ the covariance matrix of process noise. 

The sigma points $\chi^{(\ell,j,i)}_{k}$ are calculated according to
\begin{align}
\chi^{(\ell,j,0)}_{k} &=  \mu^{(\ell, j)}_{k} ,\\
\chi^{(\ell,j,i)}_{k} &=  \mu^{(\ell, j)}_{k} + \left(\sqrt{(n+\kappa)\Sigma^{(\ell, j)}_{k}}\right)_i ,\\
\chi^{(\ell,j,i+n)}_{k} &=  \mu^{(\ell, j)}_{k} - \left(\sqrt{(n+\kappa)\Sigma^{(\ell, j)}_{k}}\right)_i,
\end{align}
with corresponding weights
\begin{align} 
W^{(\ell,j,0)} &= \frac{\kappa}{n+\kappa},\\
W^{(\ell,j,i)} &= W^{(\ell,j,i+n)} = \frac{\kappa}{2(n+\kappa)},
\end{align}
where $\left(\sqrt{\Sigma}\right)_i$ is the $i$-th row of the matrix square root of $P$, $n$ is the dimension of the augmented state and $\kappa$ is a design parameter with $n+\kappa \neq 0$, shifting the sigma points further outwards or inwards in the uncertainty ellipse. Subsequently, the sigma points for $i = 1,\ldots,n$ are partitioned to 
\begin{equation}
\chi^{(\ell,j,i)}_{k} = \left[\left(\mathrm{x}^{(\ell,j,i)}_{k-1}\right)^{\text{T}}, \; \left(\mathrm{v}^{(\ell,j,i)}_{k-1}\right)^{\text{T}} \right]^{\text{T}},
\label{eq:sigmapoints}
\end{equation}
with independent zero-mean Gaussian noise process $\mathrm{v}^{(\ell,j,i)}_{k-1}$.
After the noise states are added to the corresponding state components, the prediction applies the nonlinear equations of the CTRV model, hence
\begin{equation}
\mathrm{x}^{(\ell,j,i)}_{k,+} = f\left(\mathrm{x}^{(\ell,j,i)}_{k-1}\right).
\label{eq:predicted_states}
\end{equation}
In case of a CTRV model, the states components are 
\begin{equation}
{\mathrm{x}^{(\ell,j,i)}_{k-1} = \left[ x,\; y,\; v,\; \varphi,\; \omega \right]^\text{T}},
\end{equation}
with position $x$ and $y$, magnitude of the velocity $v$, orientation $\varphi$ and turn rate $\omega$.
Hence, the state transition with sampling time $T$ is given by
\begin{equation}
\mathrm{x}^{(\ell,j,i)}_{k,+} = 
\mathrm{x}^{(\ell,j,i)}_{k-1} + 
\begin{bmatrix}
\frac{v}{\omega} \left( \sin(\varphi + \omega T) - \sin(\varphi) \right) \\
\frac{v}{\omega} \left(\cos(\varphi) - \cos(\varphi + \omega T)\right) \\
0 \\
\omega T \\
0
\end{bmatrix}.
%\begin{bmatrix}
%x + \frac{v}{\omega} \left( \sin(\varphi + \omega T) - \sin(\varphi) \right) \\
%y + \frac{v}{\omega} \left(\cos(\varphi) - \cos(\varphi + \omega T)\right) \\
%v \\
%\varphi + \omega T \\
%\omega
%\end{bmatrix}.
\label{eq:ctrvturnrate}
\end{equation}

Then, the predicted GM component is $\mathcal{N} \left(\mathrm{x}_{k,+}; \hat{\mathrm{x}}^{(\ell, j)}_{k,+}, \mathrm{P}^{(\ell, j)}_{k,+}\right)$ with predicted mean and covariance
\begin{align}
\hat{\mathrm{x}}^{(\ell, j)}_{k,+} &= \sum_{i = 0}^{2n} W^{(\ell,j,i)} \, \mathrm{x}^{(\ell,j,i)}_{k,+},
\label{eq:predicted_mean}\\
\mathrm{P}^{(\ell, j)}_{k,+} &= \sum_{i = 0}^{2n} W^{(\ell,j,i)} \, \left(\mathrm{x}^{(\ell,j,i)}_{k,+} - \hat{\mathrm{x}}^{(\ell, j)}_{k,+} \right) \left(\mathrm{x}^{(\ell,j,i)}_{k,+} - \hat{\mathrm{x}}^{(\ell, j)}_{k,+} \right)^\text{T}.
\label{eq:predicted_covariance}
\end{align}
For more details as well as the equations for the update step using a nonlinear measurement model, see \cite{Vo2008}.
%Since the equations for the LMB filter are the same as for the cardinality balanced multi-target multi-Bernoulli filter, see \cite{Vo2008}, for more details and the equations for the update step using a nonlinear measurement model.

\subsection{Approximation of a Highly Accurate Digital Map}
\label{ssec:approximateddigialmap}

The road map used in the test vehicles for autonomous driving at Ulm University \cite{Kunz2015}, is a highly precise measured digital map recorded with a Differential Global Positioning System (DGPS) mounted on the vehicles. 
For each lane on the test route, a reference line was recorded and represented as line string, i.e. a huge number of individual equidistant points. 
This format of the road is not very suitable, since single points contain less information about the road lane, e.g. width and direction of lanes or relation between multiple lanes, especially at intersections.
Therefore, in this paper the digital map is approximated by related rectangles describing road lanes. 
For the rectangles calculation, the points on the reference lines are reduced using the iterative end-point fit algorithm \cite{Ramer1972}. 
Then a set of rectangles $\mathcal{R}$ are fitted to approximate the reference lines, whereby each rectangle is represented by a center position, width, length, orientation and an unique identifier.
The curvier the road course, the more rectangles are needed for the approximated line.
The rectangles length is given by the results of the iterative end-point fit algorithm, the width is a fix parameter.
Improving the road map approximation, the rectangle width can be adapted to the actual width of the track using an offline road map with recorded lane width or an online lane detection procedure. 
Since the calculated rectangles represent a road course, the orientation of rectangles describe the direction of the road course. 
Further on, each rectangle contains information of following rectangles, which simplifies the association of rectangles to a specific road lane.

\begin{figure}[tbp]
    \vspace{1.5mm}
    \centering
    \setlength{\figH}{2.5cm}
    \setlength{\figW}{\columnwidth}
    \resizebox{\columnwidth}{!}{
%\tikzsetnextfilename{#1}
   	\small % decrease font size for tikz pictures
	\input{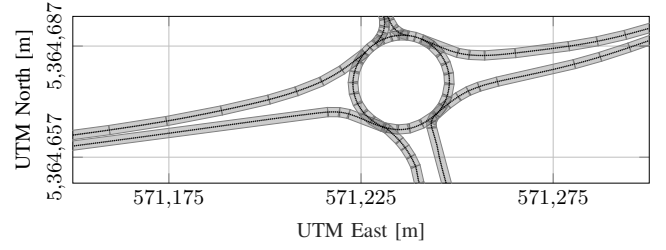}}
    \caption{Representation of a roundabout in the road map. Recorded reference lines (black dotted lines) and approximation using rectangles (gray rectangles).}
    \label{fig:digitalmap}
\end{figure}

Figure \ref{fig:digitalmap} shows recorded reference lines (black dotted lines) and  the approximated road map using rectangles representing the current road geometry (gray rectangles). 
A big advantage of this map format is, objects can be easily assigned to rectangles. 
Further, the orientation of rectangles can be used to adapt the turn rate of an object during the prediction step.

\subsection{LMB Filter Modeling Interaction Between Dynamic Tracks and Road Map Information}

As described in Section \ref{ssec:gmlmbfilter}, the prediction applies the CTRV model describing the vehicle dynamics. 
This model assumes a constant velocity and a constant turn rate, so during the prediction of a state, the velocity and turn rate remains unchanged.
In many scenarios these assumptions are not fulfilled.
This section presents a prediction containing a correction of the wrong assumptions by integrating the velocity of interacting vehicles and the orientation of the road course.

\subsubsection{Interaction Between Tracks}

Since vehicles interact primarily with the vehicle ahead, e.g. in dense traffic situations vehicles adjust their velocity according to leading vehicle, the assumption of constant velocity is wrong.
Realizing this behavior for tracks, it has to be known which tracks are moving one behind other.
Then the interactive acceleration (\ref{eq:vidm}) can be used to adapt the velocity component of a track in the prediction step using the following algorithm.

The GM (\ref{eq:gaussian_mixture}) represents the spatial distribution of a track with label $\ell$, so each component of the GM has to predict considering interacting tracks.
Firstly, for each GM component $\mathcal{N} \left(\mathrm{x}; \hat{\mathrm{x}}^{(\ell, j)},\mathrm{P}^{(\ell, j)}\right)$ the subset of rectangles $R \subseteq \mathcal{R}$ where the component is located are determined (a track can be located in more than one rectangle, since rectangles partially overlap at intersections).
Secondly, the algorithm checks for all other existing tracks with a specific existence probability, whether a track $\ell_n \neq \ell_m$ is located in front of the currently considered GM component.
A potential interacting track is located in the same rectangle as the GM component or in one of the following rectangles.
Thirdly, the predicted mean and covariance of the GM component is calculated using equations (\ref{eq:augmentedmean}) - (\ref{eq:predicted_covariance}).
If an interacting track $\ell_n$ is found, for each sigma point generated from the GM component, the velocity is adjusted using the IDM (Sec. \ref{ssec:idm}).
So the adapted velocity component of sigma point (\ref{eq:ctrvturnrate}) is
\begin{equation}
    v^{(\ell, j, i)} = \dot{v}^{(\ell, j,i)} \cdot T = -a \left( \dfrac{s^{\ast}(v^{(\ell, j, i)},\Delta v^{(\ell, j, i)})}{s^{(\ell, j, i)}}\right)^2,
\end{equation}
where $v^{(\ell, j, i)}$ is the velocity of the sigma point, $\Delta v^{(\ell, j, i)}$ the relative velocity between the sigma point and estimated mean of the interacting tracks and $s^{(\ell, j, i)}$ the distance of the sigma point to the estimated mean value of the interacted track.
As a result, this prediction compensates the error due to the wrong model assumptions of constant velocity in case that two tracks interact with each other.

\subsubsection{Dependency to Road Map}

Vehicle dynamics is limited to the road course, e.g. vehicles adjust their orientation regarding to the current road course.
Especially on curvy roads or road section with changing orientation, e.g. roundabouts, the assumption of constant turn rate is not fulfilled.
Counteracting this, the following approach uses the orientation of the road map to correct wrong orientations of tracks during the prediction.
Firstly, for each GM component $\mathcal{N} \left(\mathrm{x}; \hat{\mathrm{x}}^{(\ell, j)},\mathrm{P}^{(\ell, j)}\right)$ the subset of rectangles $R \subseteq \mathcal{R}$, where the component is located, are determined.
Secondly, during the prediction of mean and covariance, the orientation of the rectangle is used to correct the turn rate of all sigma points (\ref{eq:sigmapoints}).
For each sigma point the difference between the orientation of the rectangle $\varphi_R^{(i)}$ and the sigma point $\varphi^{(\ell, j, i)}$ is calculated, hence
\begin{equation}
\Delta \varphi^{(\ell, j, i)} = \varphi_R^{(i)} - \varphi^{(\ell, j, i)}.
\end{equation}
Finally, the new turn rate is calculated by
\begin{equation}
\omega^{(\ell, j, i)} = \Delta \varphi^{(\ell, j, i)} \cdot \dfrac{s}{v^{(\ell, j, i)}},
\label{eq:turnrate}
\end{equation}
where $s$ is the distance of the sigma point to the longitudinal border of the rectangle and $v^{(\ell, j, i)}$ its velocity.
So, the last part of equation (\ref{eq:turnrate}) can be interpreted as the time how long the sigma point will be located in the current rectangle.
As a result, using the turn rate (\ref{eq:turnrate}) the sigma point orientation converges to the orientation of the rectangle.

At intersections, a GM component is located in more than one rectangle, since there are different possibilities following the road course.
Considering all possible direction, the prediction splits up the GM component regarding the number of rectangles, i.e. it divides the weight $w^{(\ell,j)}$ into several weights uniformly distributed and predicts the multiplied components $\mathcal{N} \left(\mathrm{x}; \hat{\mathrm{x}}^{(\ell, j)},\mathrm{P}^{(\ell, j)}\right)$ with respect to the respective rectangles.
So using multiple GM components facilitate to consider various hypotheses in the prediction.

\section{Results}
\label{sec:results}
\begin{figure}[thpb]
    \vspace{1.5mm}
    \centering
    \resizebox{\columnwidth}{!}{
    \small
    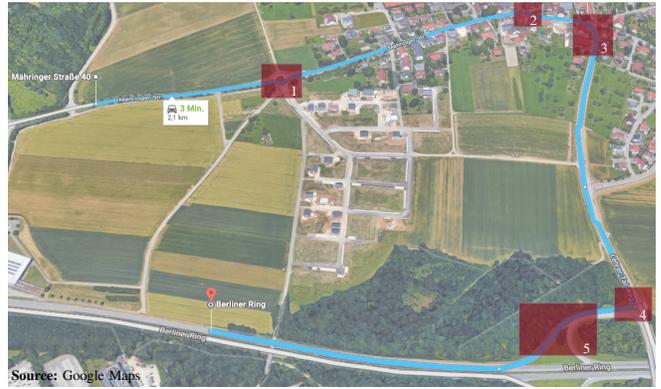}%{\includegraphics{img/Testrunde.png}}%{\includesvg{img/Testrunde}}
    \caption{Test route over $2.1\, \text{km}$ containing five challenging scenarios: (1) roundabout, (2) an urban intersection, (3) a long right turn, (4) a rural intersection and (5) an s-shaped road course.}
    \label{fig:testroute}
\end{figure}

In this chapter, results for the tracking system integrating interaction between objects and road map information are presented and compared to the standard LMB filter.
The algorithms are implemented in C++ and applied to different experimental scenarios using the test vehicles of Ulm University.
The test vehicles is equipped with a long range radar and a gray scale camera to detect and track traffic participants over time.
The first part of the evaluation examines the accuracy of estimation using a single-object tracking scenario.
For this purpose, the ego vehicle is tracking a leading vehicle on test route with rural and urban route sections (Fig. \ref{fig:testroute}).
The second part evaluates the multi-object aspect on a urban scenario with heave traffic and interaction between vehicles.

\subsection{Practical Implementation}

Modeling the object dynamic during the prediction, a CTRV model is used.
The parameters for the IDM in (\ref{eq:sidm}) are determined experimentally and in accordance with \cite{Treiber2000} given by $T = 1.6\,\text{s}$, $a = 0.73\,\text{m}/\text{s}^2$ and $b = 1.67\,\text{m}/\text{s}^2$.
Since the assumptions of this model are not fulfilled at all times, process noise is added to an object's state.
The normalized values are $5\,\text{m/s}^2$ for the velocity, $0.1\,\text{rad/s}^2$ for the turn rate.
The persistance probability of a track is given by $0.99$.
Further, in the measurement models the detection probability of the radar is set to $0.85$ and of the camera to $0.75$, the clutter intensity of the radar to $1e^{-5}$ and of the camera to $0.01$.
Tracks are extracted with an existence probability greater than $0.2$ and are pruned with a probability less than $0.01$.

\begin{figure*}[ht]
    \vspace{1.5mm}
    \centering
    \setlength{\figH}{7.5cm}
    \setlength{\figW}{\textwidth}
    \resizebox{\textwidth}{!}{
%\tikzsetnextfilename{#1}
   	\small % decrease font size for tikz pictures
\begin{tikzpicture}

\begin{axis}[%
width=0.426\figW,
height=0.208\figH,
at={(0\figW,0.781\figH)},
scale only axis,
unbounded coords=jump,
xmin=0,
xmax=150,
xlabel={Time [s]},
xmajorgrids,
ymin=-10,
ymax=20,
ylabel={$e_{x}\text{[m]}$},
ymajorgrids,
axis background/.style={fill=white}
]
\addplot [color=red,solid,line width=1.0pt,forget plot]
  table[]{img/tikz/Testrunde-1.tsv};
\addplot [color=blue,dashed,line width=1.0pt,forget plot]
  table[]{img/tikz/Testrunde-2.tsv};
\end{axis}

\begin{axis}[%
width=0.426\figW,
height=0.208\figH,
at={(0.574\figW,0.781\figH)},
scale only axis,
xmin=0,
xmax=150,
xlabel={Time [s]},
xmajorgrids,
ymin=-6,
ymax=4,
ytick={-5, -2.5, 0, 2.5, 5},
ylabel={$e_{y}\text{[m]}$},
ymajorgrids,
axis background/.style={fill=white}
]
\addplot [color=red,solid,line width=1.0pt,forget plot]
  table[]{img/tikz/Testrunde-3.tsv};
\addplot [color=blue,dashed,line width=1.0pt,forget plot]
  table[]{img/tikz/Testrunde-4.tsv};
\end{axis}

\begin{axis}[%
width=0.426\figW,
height=0.208\figH,
at={(0\figW,0.39\figH)},
scale only axis,
xmin=0,
xmax=150,
xlabel={Time [s]},
xmajorgrids,
ymin=-0.75,
ymax=0.75,
ylabel={$e_{\varphi}\text{[rad]}$},
ymajorgrids,
axis background/.style={fill=white}
]
\addplot [color=red,solid,line width=1.0pt,forget plot]
  table[]{img/tikz/Testrunde-5.tsv};
\addplot [color=blue,dashed,line width=1.0pt,forget plot]
  table[]{img/tikz/Testrunde-6.tsv};
\end{axis}

\begin{axis}[%
width=0.426\figW,
height=0.208\figH,
at={(0.574\figW,0.39\figH)},
scale only axis,
unbounded coords=jump,
xmin=0,
xmax=150,
xlabel={Time [s]},
xmajorgrids,
ymin=-5,
ymax=15,
ylabel={$e_{v}\text{[m/s]}$},
ymajorgrids,
axis background/.style={fill=white}
]
\addplot [color=red,solid,line width=1.0pt,forget plot]
  table[]{img/tikz/Testrunde-7.tsv};
\addplot [color=blue,dashed,line width=1.0pt,forget plot]
  table[]{img/tikz/Testrunde-8.tsv};
\end{axis}

\begin{axis}[%
width=0.426\figW,
height=0.208\figH,
at={(0\figW,0\figH)},
scale only axis,
unbounded coords=jump,
xmin=0,
xmax=150,
xlabel={Time [s]},
xmajorgrids,
ymin=-0.5,
ymax=0.5,
ytick={-0.25, 0, 0.25},
ylabel={$e_{\omega}\text{[rad/s]}$},
ymajorgrids,
axis background/.style={fill=white}
]
\addplot [color=red,solid,line width=1.0pt,forget plot]
  table[]{img/tikz/Testrunde-9.tsv};
\addplot [color=blue,dashed,line width=1.0pt,forget plot]
  table[]{img/tikz/Testrunde-10.tsv};
\end{axis}

\begin{axis}[%
width=0.426\figW,
height=0.208\figH,
at={(0.574\figW,0\figH)},
scale only axis,
xmin=0,
xmax=150,
xlabel={Time [s]},
xmajorgrids,
ymin=0,
ymax=5,
ylabel={Label $\ell$},
ymajorgrids,
axis background/.style={fill=white}
]
\addplot [color=red,solid,line width=1.0pt,forget plot]
  table[]{img/tikz/Testrunde-11.tsv};
\addplot [color=blue,dashed,line width=1.0pt,forget plot]
  table[]{img/tikz/Testrunde-12.tsv};
\end{axis}
\end{tikzpicture}%
}
    \caption{Comparison of resulting estimation errors $e$ using the standard LMB filter (dashed blue) and the LMB filter modeling interacting objects and road map information (solid red).}
    \label{fig:evaltestrunde}
\end{figure*}
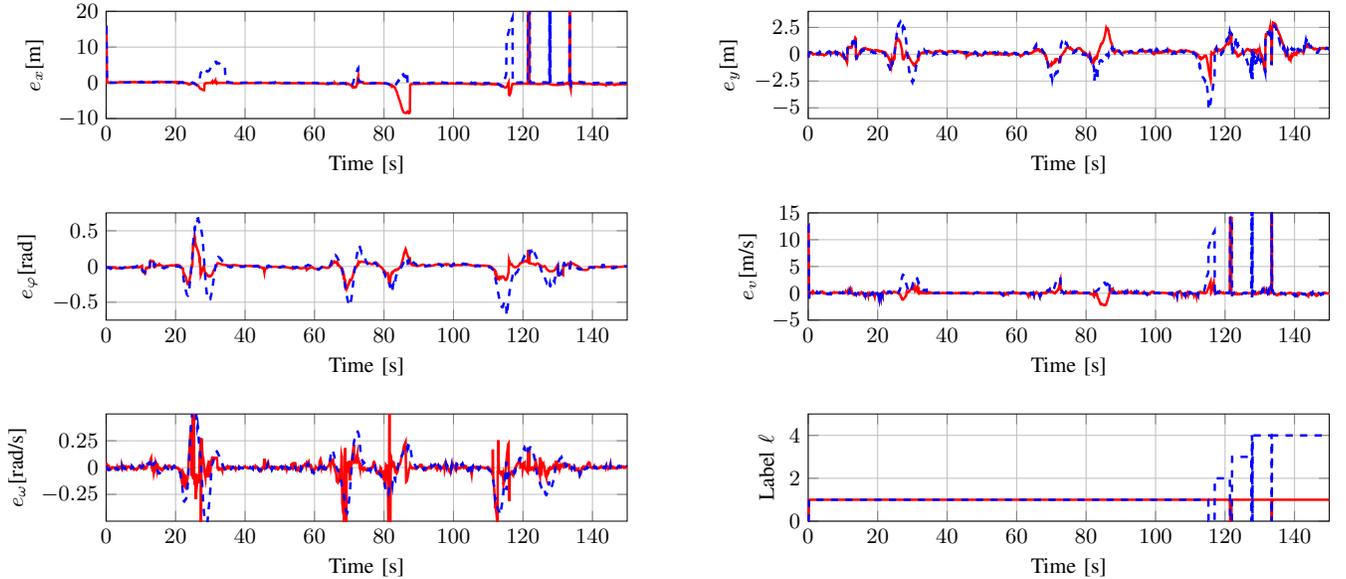 

\subsection{Single-Object Accuracy}

Figure \ref{fig:testroute} shows the $2.1\, \text{km}$ test route located in Ulm containing different challenging situations for the tracking system.
There are a roundabout (1), a long right turn (3) and an s-shaped road course (5), where assumptions of the vehicle dynamic are violated, because changes occur in the turn rate.
Further, there are two intersections, an urban crossroad (2) and a rural crossroad (4), where it is not clear what direction an object turns.
For this scenario, two test vehicles drive along the blue marked route.
During this scenario, interaction of the reference vehicle with a vehicle ahead occurs, as well as occlusion of the reference vehicle by another road users.
Ground truth data of the reference vehicle are provided by a DGPS.

Figure \ref{fig:evaltestrunde} visualizes the resulting estimation errors for all state components as well as the error for the estimated label.
Most of the time, the LMB filter modeling interacting objects and road map information outperforms the standard LMB filter, since the error in respective state components is smaller.
The label error is interpreted as follow:
If the reference vehicle is not tracked, because the existence probability is too low, the value is zero.
If a new track is initialized, the value increases by one.
Best result would be a constant value one, then the track label of the object regarding to the reference vehicle is estimated correctly over the entire scenario.

Table \ref{ta:rmse} lists root mean squared error (RMSE) values for the scenario.
Obviously, for all state vector components, estimation results using the LMB filter modeling interaction and road map information are more precise.
Compared to the standard LMB filter, the accuracy is improved by $23 - 55\%$.

\begin{table}[tpb]
     \caption{RMSE values using the LMB filter modeling interaction and road map information and the standard LMB filter.}
     \centering
     \begin{tabular}{c c c c }
         \hlinewd{1pt}
         States & Interacting LMB & Standard LMB &  Improvement [\%] \\ 
         \hline
         $x$ [m] & $1.95$ &  $2.80$ & $30.62$ \\ 
         $y$ [m] & $0.63$ &  $0.96$ & $33.70$ \\ 
         $\varphi$ [$^\circ$] & $4.10$ &  $9.09$ & $54.94$ \\ 
         $v$ [m/s] & $1.10$ &  $1.86$ & $40.86$ \\ 
         $\omega$ [$^\circ$/s] & $5.72$ &  $7.48$ & $23.63$ \\ 
         \hlinewd{1pt} 
     \end{tabular} 
     \label{ta:rmse}
\end{table}

\begin{figure}[thpb]
    \vspace{1.5mm}
    \centering
    \resizebox{\columnwidth}{!}{\includegraphics{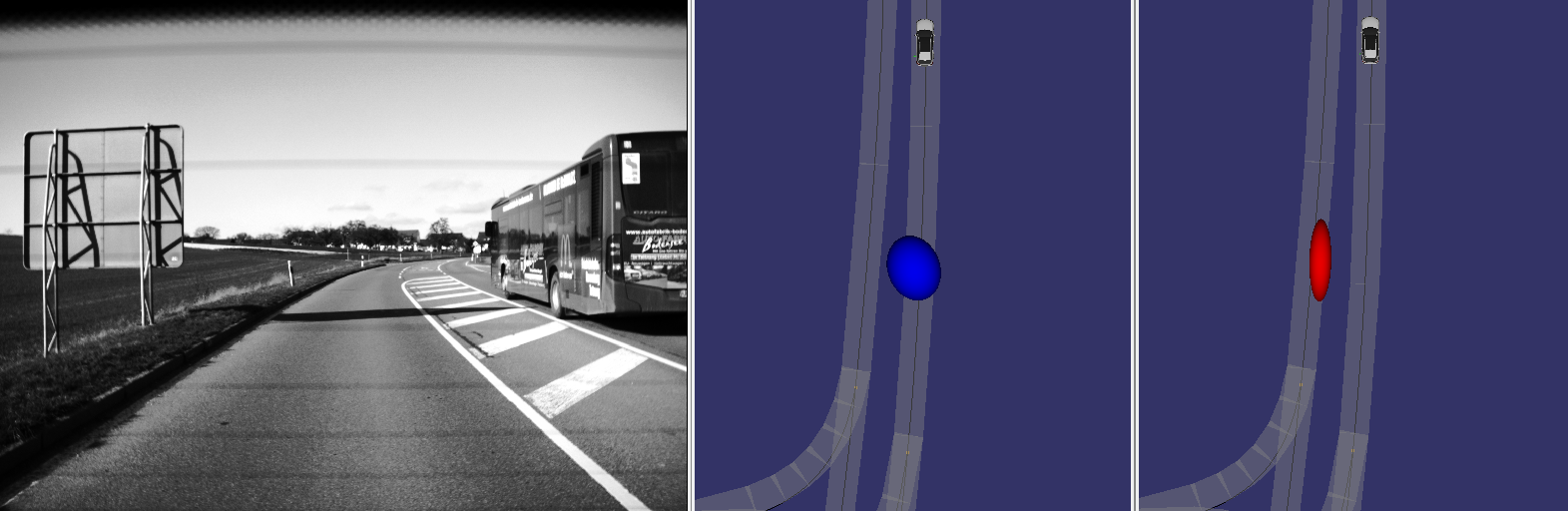}}
    \caption{On the left, the view of the rear camera showing a bus on the opposite lane. In the middle and right, mean and covariance of estimated object which belongs the bus. The blue ellipse belongs to the standard tracking system, the red to the improved system integrating road map information (gray rectangles).}
    \label{fig:uncertaintyellipse}
\end{figure}

Another major benefit of the presented approach integrating road map information to adapt the turn rate, is a more realistic estimation of the covariances.
By adjusting sigma points (\ref{eq:ctrvturnrate}) using road map information, not only the mean (\ref{eq:predicted_mean}) is affected, but also the covariance (\ref{eq:predicted_covariance}) implicitly.
Figure \ref{fig:uncertaintyellipse} illustrates this fact. 
With the standard LMB tracker, the oncoming bus is estimated on the wrong lane behind the ego vehicle with a circular expanding covariance.
Using road map information, the mean of the track is located on the correct lane and the covariance realistically expands towards the course of road.

\subsection{Multi-Object Performance}

\begin{figure}[tpb]
    \vspace{1.5mm}
    \centering  
    \subfigure[\label{subfig:orientation}]{\resizebox{\columnwidth}{!}{\includegraphics{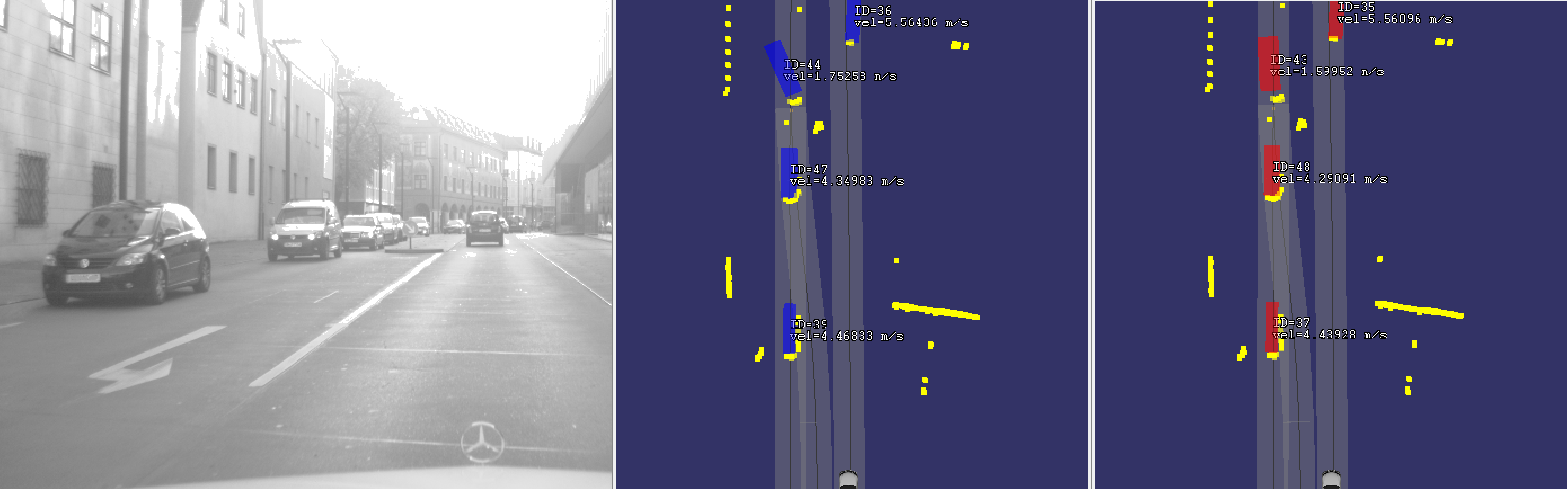}}} 
    \subfigure[\label{subfig:velocity}]{\resizebox{\columnwidth}{!}{\includegraphics{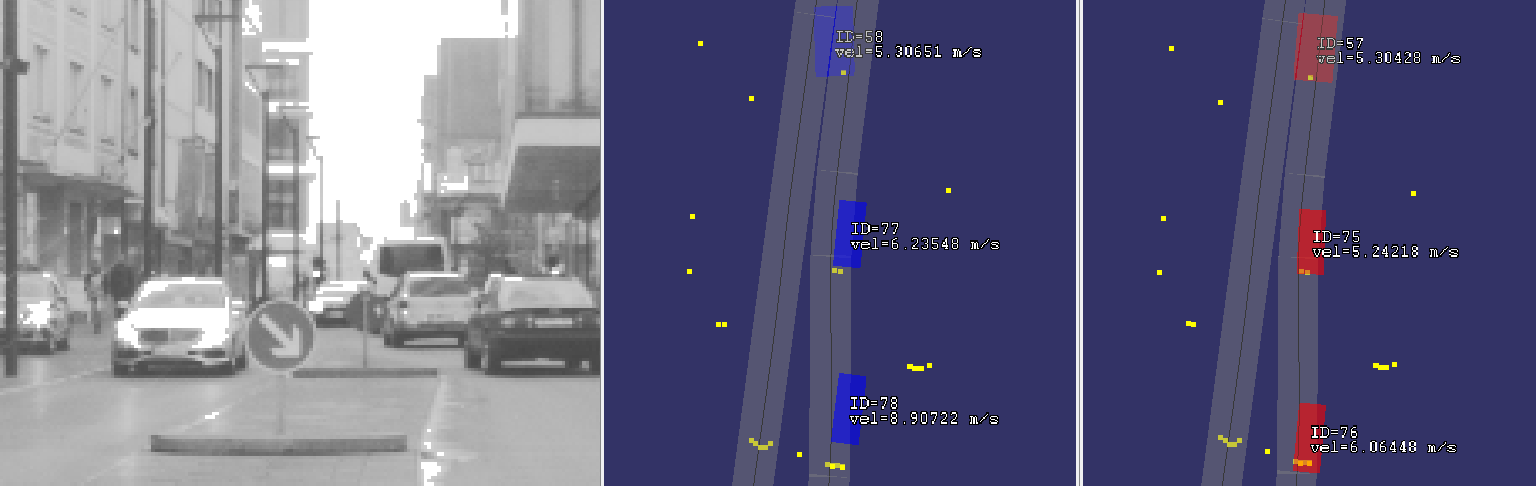}}} 
    \caption{On the left, the view of the front camera showing an urban scenario with dense traffic. In the middle, estimated objects of the standard LMB tracker (blue boxes), and on the right, the results of the LMB tracker using interacting objects and road map information (red boxes). The laser measurements (yellow points) visualize position and shape of vehicles.} 
    \label{fig:multiobjecttracking}
\end{figure}

Since no ground truth data is available from urban scenarios with dense traffic, this part of the evaluation is qualitative.
Figure \ref{fig:multiobjecttracking} shows two scenarios in an typically urban area with dense traffic.
In both scenarios, occlusion between vehicles occurs, so the existing vehicles do not generate measurements all the time.
Especially in such situations, the prediction affects the tracking results, hence modeling interaction between objects and road map information can be very useful.
Evaluating the estimated objects, the camera view is used to understand the scene, and a precise laserscanner to get an idea of the vehicles' position and shape.

Figure \ref{subfig:orientation} shows a situation of three oncoming vehicles.
Here, particularly the integration of the road map information improves the tracking results.
The standard LMB filter estimates the orientation of track $44$ (blue box) clearly incorrect.
Using the road map information, the estimation of the same track (red box with label $43$) is much more accurate.
Since the velocity of the three objects in a row is low and the safety distance calculated by the IDM is maintained, the difference in the velocity estimation is small.
Figure \ref{subfig:velocity} visualizes a situation with three strongly interacting vehicles driving ahead the ego vehicle.
Here, again integrating road map information results in a better estimated orientation, so all tracked objects (red boxes) are more orientated towards the course of road.
In this scenario, the effect of modeling interaction is clearly visible.
Considering objects $77$ and $78$ estimated by the standard LMB filter (blue boxes), the estimated velocity is too high.
As a result, the error between estimated position and respective laser points is clearly visible, especially looking at object $78$.
Modeling interaction between these three objects improves the corresponding estimated velocities (red boxes with label $75$ and $76$).
Thus, the position error can be significantly reduced.

\section{Conclusion}
\label{sec:conclusion}
This contribution has proposed an approach for multi-object tracking modeling interaction between estimated objects as well as integrating road map information using an approximation of a highly precise digital map.
Especially in situations with dense traffic or curvy road course, the model assumptions describing the vehicle dynamics are not fulfilled.
For this reason, the presented approach has integrated interacting objects and road map information into a standard vehicle dynamic model adapting the prediction step of a tracking system.
On scenarios with dense traffic and interacting objects, the estimated velocity component is improved using the Intelligent Driver Model.
Further, an approximated road map described by connected rectangles is used to adapt the turn rate component.
The evaluation has shown, modeling interaction between objects improves tracking results, particularly in scenarios with dense traffic, e.g. urban traffic, since the dependency between vehicles is modeled in the prediction.
Further, in situations with curvy road lanes, e.g. round abouts or intersections, the results are much better since the road map information is very useful.
As a result, using road map information leads to a significantly accurate estimation of the orientation, and as a result existing tracks are lost less often.

\section*{Acknowledgment}
This project was funded within the Priority Program "Cooperatively Interacting Automobiles" of the German Research Foundation (DFG).

%%%%%%%%%%%%%%%%%%%%%%%%%%%%%%%%%%%%%%%%%%%%%%%%%%%%%%%%%%%%%%%%%%%%%%%%%%%%%%%%

\bibliographystyle{IEEEtran}
\bibliography{IEEEabrv,references}

\begin{thebibliography}{10}
\providecommand{\url}[1]{#1}
\csname url@rmstyle\endcsname
\providecommand{\newblock}{\relax}
\providecommand{\bibinfo}[2]{#2}
\providecommand\BIBentrySTDinterwordspacing{\spaceskip=0pt\relax}
\providecommand\BIBentryALTinterwordstretchfactor{4}
\providecommand\BIBentryALTinterwordspacing{\spaceskip=\fontdimen2\font plus
\BIBentryALTinterwordstretchfactor\fontdimen3\font minus
  \fontdimen4\font\relax}
\providecommand\BIBforeignlanguage[2]{{%
\expandafter\ifx\csname l@#1\endcsname\relax
\typeout{** WARNING: IEEEtran.bst: No hyphenation pattern has been}%
\typeout{** loaded for the language `#1'. Using the pattern for}%
\typeout{** the default language instead.}%
\else
\language=\csname l@#1\endcsname
\fi
#2}}

\bibitem{Bar-Shalom1988}
Y.~Bar-Shalom and T.~Fortmann, \emph{{Tracking and Data Association}}.\hskip
  1em plus 0.5em minus 0.4em\relax Academic Press, Inc., 1988.

\bibitem{Reid1979}
D.~Reid, ``{An Algorithm for Tracking Multiple Targets},'' \emph{IEEE
  Transactions on Automatic Control}, vol.~24, no.~6, pp. 843--854, 12 1979.

\bibitem{Mahler2007}
R.~Mahler, \emph{{Statistical Multisource-Multitarget Information
  Fusion}}.\hskip 1em plus 0.5em minus 0.4em\relax Artech House Inc., Norwood,
  2007.

\bibitem{Vo2013}
B.-T. Vo and B.-N. Vo, ``{Labeled Random Finite Sets and Multi-Object Conjugate
  Priors},'' \emph{IEEE Transactions on Signal Processing}, vol.~61, no.~13,
  pp. 3460--3475, 2013.

\bibitem{Reuter2014}
S.~Reuter, B.-T. Vo, B.-N. Vo, and K.~Dietmayer, ``{The Labeled
  Multi-{B}ernoulli Filter},'' \emph{IEEE Transactions on Signal Processing},
  vol.~62, no.~12, pp. 3246 -- 3260, 2014.

\bibitem{Reuter2014b}
S.~Reuter, ``{Multi-Object Tracking Using Random Finite Sets},'' Ph.D.
  dissertation, Ulm University, 2014.

\bibitem{Luber2010}
M.~Luber, J.~A. Stork, G.~D. Tipaldi, and K.~O. Arras, ``{People Tracking with
  Human Motion Predictions from Social Forces},'' in \emph{Proceedings
  International Conference on Robotics and Automation}.\hskip 1em plus 0.5em
  minus 0.4em\relax IEEE, 2010.

\bibitem{Helbing1995}
D.~Helbing and P.~Molnar, ``{Social Force Model for Pedestrian Dynamics},''
  \emph{Physical Review E}, vol.~51, pp. 4282 -- 4286, 1995.

\bibitem{Feng2017}
P.~Feng, W.~Wang, S.~Dlay, S.~M. Naqvi, and J.~Chambers, ``{Social Force
  Model-Based MCMC-OCSVM Particle PHD Filter for Multiple Human Tracking},''
  \emph{IEEE Transactions on Multimedia}, vol.~19, no.~4, pp. 725 -- 739, April
  2017.

\bibitem{Mahler2003}
R.~Mahler, ``{Multitarget Bayes Filtering via First-Order Multitarget
  Moments},'' \emph{IEEE Transactions on Aerospace and Electronic Systems},
  vol.~39, no.~4, pp. 1152--1178, 10 2003.

\bibitem{Ding2017}
R.~Ding, M.~Yu, H.~Oh, and W.-H. Chen, ``{New Multiple-Target Tracking Strategy
  Using Domain Knowledge and Optimization},'' in \emph{Transactions on Systems,
  Man and Cybernetics}, vol.~47, no.~4.\hskip 1em plus 0.5em minus 0.4em\relax
  IEEE, April 2017.

\bibitem{Helbing1998}
D.~Helbing and B.~Tilch, ``{Generalized Force Model of Traffic Dynamics},''
  \emph{Physical Review E}, vol.~58, pp. 133 -- 138, 1998.

\bibitem{Helly1959}
W.~Helly, ``{Simulation of Bottlenecks in Single Lane Traffic Flow},'' in
  \emph{Proceedings Theory Traffic Flow}, April 1959, pp. 207 -- 238.

\bibitem{Treiber2000}
M.~Treiber, A.~Hennecke, and D.~Helbing, ``{Congested Traffic States in
  Empirical Observations and Microscopic Simulations},'' \emph{Physical Review
  E}, vol.~62, no.~2, pp. 1805 -- 1824, August 2000.

\bibitem{Song2017}
D.~Song, R.~Tharmarasa, T.~Kirubarajan, and X.~N. Fernando, ``{Multi-Vehicle
  Tracking With Road Maps and Car-Following Models},'' \emph{IEEE Transactions
  on Intelligent Transportation Systems}, vol.~PP, no.~99, pp. 1 -- 12, 2017.

\bibitem{Kalman1960}
R.~E. Kalman, ``{A New Approach to Linear Filtering and Prediction Problems},''
  \emph{Transactions of the ASME - Journal of Basic Engineering}, vol.~82, pp.
  35--45, 1960.

\bibitem{Julier2004}
S.~Julier and J.~Uhlmann, ``{Unscented Filtering and Nonlinear Estimation},''
  \emph{Proceedings of the IEEE}, vol.~92, no.~3, pp. 401--422, 3 2004.

\bibitem{Kunz2015}
F.~Kunz, D.~Nuss, J.~Wiest, \emph{et~al.}, ``{Autonomous Driving at Ulm
  University: A Modular, Robust, and Sensor-Independent Fusion Approach},'' in
  \emph{Intelligent Vehicles Symposium}, 2015, pp. 666--673.

\bibitem{Schubert2008}
R.~Schubert, E.~Richter, and G.~Wanielik, ``{Comparison and Evaluation of
  Advanced Motion Models for Vehicle Tracking},'' in \emph{11th International
  Conference on Information Fusion}.\hskip 1em plus 0.5em minus 0.4em\relax
  IEEE, 2008.

\bibitem{Vo2009}
B.-T. Vo, B.-N. Vo, and A.~Cantoni, ``{The Cardinality Balanced Multi-Target
  Multi-{Bernoulli} Filter and Its Implementations},'' \emph{IEEE Transactions
  on Signal Processing}, vol.~57, no.~2, pp. 409--423, 2009.

\bibitem{Vo2008}
B.~T. Vo, ``{Random Finite Sets in Multi-Object Filtering},'' Ph.D.
  dissertation, The University of Western Australia, 2008.

\bibitem{Ramer1972}
U.~Ramer, ``{An Iterative Procedure for the Polygonal Approximation of Plane
  Curves},'' \emph{Computer Graphics and Image Processing}, vol.~1, no.~3, pp.
  244 -- 256, 1972.

\end{thebibliography}

\end{document}